# CLP($\mathcal{BN}$): Constraint Logic Programming for Probabilistic Knowledge

**Vítor Santos Costa**
COPPE/Sistemas
UFRJ, Brasil

**David Page and Maleeha Qazi**
Department of Biostatistics
and Medical Informatics
University of Wisconsin-Madison, USA

**James Cussens**
Department of Computer Science
University of York, UK

## Abstract

In Datalog, missing values are represented by Skolem constants. More generally, in logic programming missing values, or existentially-quantified variables, are represented by terms built from Skolem functors. In an analogy to probabilistic relational models (PRMs), we wish to represent the joint probability distribution over missing values in a database or logic program using a Bayesian network. This paper presents an extension of logic programs that makes it possible to specify a joint probability distribution over terms built from Skolem functors in the program. Our extension is based on constraint logic programming (CLP), so we call the extended language CLP($\mathcal{BN}$). We show that CLP($\mathcal{BN}$) subsumes PRMs; this greater expressivity carries both advantages and disadvantages for CLP($\mathcal{BN}$). We also show that algorithms from inductive logic programming (ILP) can be used with only minor modification to learn CLP($\mathcal{BN}$) programs. An implementation of CLP($\mathcal{BN}$) is publicly available as part of YAP Prolog at http://www.cos.ufrj.br/˜vitor/Yap/clpbn.

## 1 Introduction

A probabilistic relational model (PRM) [4] uses a Bayesian network to represent the joint probability distribution over fields in a relational database. The Bayes net can be used to make inferences about missing values in the database. In Datalog, missing values are represented by Skolem constants; more generally, in logic programming missing values, or existentially-quantified variables, are represented by terms built from Skolem functors. In analogy to PRMs, can a Bayesian network be used to represent the joint probability distribution over terms constructed from the Skolem functors in a logic program? We extend the language of logic programs to make this possible. Our extension is based on constraint logic programming (CLP), so we call the extended language CLP($\mathcal{BN}$). We show that any PRM can be represented as a CLP($\mathcal{BN}$) program.

Our work in CLP($\mathcal{BN}$) has been motivated by our interest in multi-relational data mining, and more specifically in inductive logic programming (ILP). Because CLP($\mathcal{BN}$) programs are a kind of logic program, we can use existing ILP systems to learn them, with only simple modifications to the ILP systems. Induction of clauses can be seen as model generation, and parameter fitting can be seen as generating the CPTs for the constraint of a clause. We show that the ILP system ALEPH is able to learn CLP($\mathcal{BN}$) programs.

## 2 CLP($\mathcal{BN}$) by Example

We shall use the school database scheme originally used to explain Probabilistic Relational Models [4] (PRMs) to guide us through CLP($\mathcal{BN}$). We chose this example because it stems from a familiar background and because it illustrates how CLP($\mathcal{BN}$) relates to PRMs. Figure 1 presents the database scheme and the connection between random variables. There are four relations, describing professors, students, courses, and registrations. Field names in italics correspond to random variables.

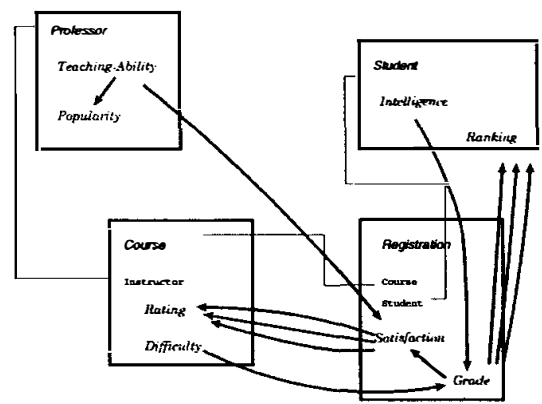

Figure 1: The School Database

Random variables may depend on other random variables, inducing the dependencies shown in Figure 1. In a nutshell, a professor's ability, a course's difficulty, and a student's intelligence do not depend on any other factors explicitly represented in the database. A professor's popularity depends only on his/her ability. A registration's grade depends on the course's difficulty, and on the student's intelligence. A student's ranking depends on *all* the grades he/she had, and a course's rating depends on the satisfaction of every student who attended the course.

One possible representation would be to use Skolem functions to represent random attributes. In the simplest case, professor ability, we would write:

```
ability(jim,skA(jim)),
```

where $skA$(jim) is a Skolem function of jim. Unfortunately, our assertion is not very illuminating. We would also like to represent the probabilities for the different cases of ability, that is, we would like to write:

$$\text{ability(jim,skA(jim))} \land$$
$$P(skA(\text{jim}) = \text{h}) = 0.7 \land P(\text{skA(jim)} = \text{l}) = 0.3$$

We would further like to use special inference rules for probabilities. Logic programming systems have used constraints to address similar problems. Logical variables are said to be *constrained* if they are bound to one or more constraints. Constraints are kept in a separate *store* and can be updated as execution proceeds (ie, if we receive new evidence on a variable). Unifying a term with a constrained variable invokes a specialised *solver*. The solver is also activated before presenting the answer to a query. In constraint notation, we could say:

$$\exists X, \text{ ability(jim,X)} \land$$
$$\{X = skA(\text{jim}) \land P(X = \text{h}) = 0.7 \land P(X = \text{l}) = 0.3\}$$

The curly brackets surround the constraint store: they say that $X$ must take the value of the function $skA$(jim), and that it has two possible values with complementary probabilities. CLP($\mathcal{BN}$) programs manipulate such constraints. We use the following syntax:

```
{Abi = a(jim) with p([h,l],[0.7,0.3],[])}
```

**CLP($\mathcal{BN}$) Programs**  A CLP($\mathcal{BN}$) is a constraint logic program that can encode Bayesian constraints. CLP($\mathcal{BN}$) programs thus consist of clauses. Our first example of a clause defines a student's intelligence in the school database:

```
intelligence(S,Int) :-
  {Int = i(S) with p([h,l],[0.7,0.3],[])}.
```

In this example we have the same information on every student's intelligence. Often, we may have different probability distributions for different students:

```
intelligence(S,Int) :-
  int_table(S, Dist),
  {Int = i(S) with p([h,l],Dist,[])}.

int_table(bob, [0.3, 0.9]) :- !.
int_table(mike, [0.8, 0.2]) :- !.
int_table( _, [0.7,0.3]).
```

Probability distributions are first class objects in our language: they can be specified at compile-time or computed from arbitrary logic programs.

**Conditional Probabilities**  Let us next consider an example of a conditional probability distribution (CPT). In Figure 1 one can observe that a registration's grade depends on the course's difficulty and on the student's intelligence. This is encoded by the following clause:

```
grade(Reg, Grade) :-
  reg(Reg,Course,Student),
  difficulty(Course,Dif),
  intelligence(Student,Int),
  {Grade = grade(Reg) with p(
   [a,b,c], [0.4,0.9,0.4,0.0
             0.4,0.1,0.4,0.1,
             0.2,0.0,0.2,0.9], [Dif,Int])}.
```

The constraint says that *Grade* is a skolem function of *Reg*, *Dif*, and *Int*. We know that *Grade* must be unique for each *Reg*, and the probability distribution for the possible values of *Grade* only depends on the Skolem variables *Dif* and *Int*. In order to keep the actual CPT small, we assume that *Dif* and *Int* each can take only two values. Note that in general, CPTs can be obtained from arbitrary logic programs which compute a number from any *(structured_term,constant)* pair.

**Execution**  The evaluation of a CLP($\mathcal{BN}$) program results in a network of constraints. In the previous example, the evaluation of

```
?- grade(r2,Grade).
```

will set up a constraint network with grade(r2) depending on dif(course) and int(student). CLP($\mathcal{BN}$) will output the marginal probability distribution on grade(r2).

One major application of Bayesian systems is conditioning on evidence. For example, if a student had a good grade we may want to find out whether he or she is likely to be intelligent:

```
?- grade(r2,a), intelligence(bob,I).
```

The user introduces evidence for r2's grade through binding the argument in the corresponding goal. In practice,

the system preprocesses the clause and adds evidence as an extra constraint on the argument.

# 3 Foundations

We next present the basic ideas of CLP($\mathcal{BN}$) more formally. For brevity, this section necessarily assumes prior knowledge of first-order logic, model theory, and resolution.

First, we remark that CLP($\mathcal{BN}$) programs are logic programs, and thus inherit the well-known properties of logic programs. We further interpret a CLP($\mathcal{BN}$) program as defining a set of probability distributions over the models of the underlying logic program. Any Skolem function $sk$ of variables $X_1, \ldots, X_n$, has an associated CPT specifying a probability distribution over the possible denotations of $sk(X_1, \ldots, X_n)$ given the values, or bindings, of $X_1, \ldots, X_n$. The CPTs associated with a clause may be thought of as a Bayes net, where each node is labelled by either a variable or a term built from a Skolem function. Figure 2 illustrates this view using a clause that relates a registration's grade to the course's difficulty and to the student's intelligence. At times we will denote a CLP($\mathcal{BN}$) clause by $C/B$, where $C$ is the logical portion and $B$ is the probabilistic portion. For a CLP($\mathcal{BN}$) program $P$, the logical portion of $P$ is simply the conjunction of the logical portions of all the clauses in $P$.

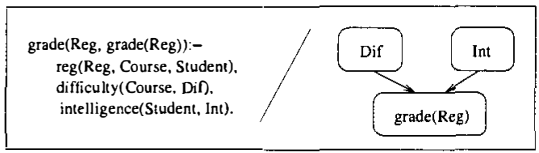

Figure 2: Pictorial representation of a grade clause.

**Detailed Syntax** The alphabet of CLP($\mathcal{BN}$) is the alphabet of logic programs. We shall take a set of functors and call these functors *Skolem functors*; *Skolem constants* are simply Skolem functors of arity 0. A Skolem term is a term whose primary functor is a Skolem functor. We assume that Skolem terms have been introduced into the program during a Skolemization process to replace the existentially-quantified variables in the program. It follows from the Skolemization process that any Skolem functor $sk$ appears in only one Skolem term, which appears in only one clause, though that Skolem term may have multiple occurrences in that one clause. Where the Skolem functor $sk$ has arity $n$, its Skolem term has the form $sk(W_1, \ldots, W_n)$, where $W_1, \ldots, W_n$ are distinct variables that also appear outside of any Skolem term in the same clause.

A CLP($\mathcal{BN}$) program in canonical form is a set of *clauses* of the form $H \leftarrow A/B$. We call $H$ the head of the clause.

$H$ is a literal and $A$ is a (possibly empty) conjunction of literals. Together they form the logical portion of the clause, $C$. The probabilistic portion, $B$, is a (possibly empty) conjunction of atoms of the form: $\{V = Sk \text{ with CPT}\}$. We shall name these atoms *constraints*. Within a constraint, we refer to $Sk$ as the Skolem term and $CPT$ as the conditional probability table. We focus on discrete variables in this paper. In this case, $CPT$ may be an unbound variable or a term or the form $\mathbf{p}(\mathbf{D}, \mathbf{T}, \mathbf{P})$. We refer to $D$ as the domain, $T$ as the table, and $P$ as the parent nodes.

A CLP($\mathcal{BN}$) constraint $B_i$ is well-formed if and only if:

1. all variables in $B_i$ appear in $C$;

2. $Sk's$ functor is unique in the program; and,

3. there is at least one substitution $\sigma$ such that $CPT\sigma = \mathbf{p}(\mathbf{D}\sigma, \mathbf{T}\sigma, \mathbf{P}\sigma)$, and **(a)** $D\sigma$ is a ground list, all members of the list are different, and no subterm of a term in the list is a Skolem term; **(b)** $P\sigma$ is a ground list, all members of the list are different, and all members of the list are Skolem terms; and **(c)** $T\sigma$ is a ground list, all members of $T\sigma$ are numbers $p$ such that $0 \le p \le 1$, and the size of $T\sigma$ is a multiple of the size of $D\sigma$.

If the probabilistic portion of a clause is empty, we also call the clause a *Prolog clause*. According to this definition, every Prolog program is a CLP($\mathcal{BN}$) program.

**Operational Semantics** A query for CLP($\mathcal{BN}$) is an ordinary Prolog query, which is a conjunction of positive literals. In logic programming, a query is answered by one or more proofs constructed through resolution. At each resolution step, terms from two different clauses may be unified. If both of the terms being unified also participate in CPTs, or Bayes net constraints, then the corresponding nodes in the Bayes net constraints must be *unified* as illustrated in Figure 3. In this way we construct a large Bayes net consisting of all the smaller Bayes nets that have been unified during resolution.

A cycle may arise in the Bayes Net if we introduce a constraint such that $Y$ is a parent of $X$, and $X$ is an ancestor of $Y$. In this case, when unifying $Y$ to an argument of the CPT constraint for $X$, $X$ would be a sub-term of the CPT constraint for $Y$: we thus can use the occur-check test to guarantee the net is acyclic.

To be rigorous in our definition of the distribution defined by a Bayes net constraint, let $C_i/B_i$, $1 \le i \le n$, be the clauses participating in the proof, where $C_i$ is the ordinary logical portion of the clause and $B_i$ is the attached Bayes net, in which each node is labelled by a term. Let $\theta$ be the answer substitution, that is, the composition of the most general unifiers used in the proof. Note that during resolution a clause may be used more than once but its variables always are renamed, or standardised apart from variables used earlier. We take each such renamed clause used in the proof to be a distinct member of $\{C_i/B_i | 1 \leq i \leq n\}$. We define the application of a substitution $\theta$ to a Bayes net as follows. For each node in the Bayes net, we apply $\theta$ to the label of that node to get a new label. If some possible values for that node (according to its CPT) are not instances of that new label, then we marginalise away those values from the CPT.

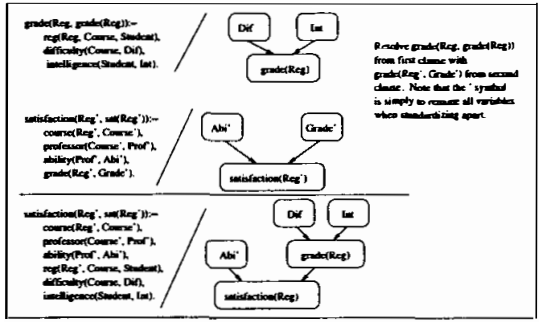

Figure 3: Resolution.

**Model-theoretic Semantics**  A CLP($\mathcal{BN}$) program denotes a probability distribution over models. We begin by defining the probability distribution over ground Skolem terms that is specified by the probabilistic portion of a CLP($\mathcal{BN}$) program. We then specify the probability distribution over models, consistent with this probability distribution over ground Skolem terms, that the full CLP($\mathcal{BN}$) program denotes.

A CLP($\mathcal{BN}$) program $P$ defines a unique joint probability distribution over ground Skolem terms as follows. Consider each ground Skolem term to be a random variable whose domain is a finite set of non-Skolem constants.[1] We now specify a Bayes net $\mathcal{BN}$ whose variables are these ground Skolem terms. Each ground Skolem term $s$ is an instance of exactly one Skolem term $t$ in the program $P$. To see this recall that, from the definition of Skolemization, any Skolem functor appears in only one term in the program $P$, and this one term appears in only one clause of $P$, though it may appear multiple times in that clause. Also from the definition of Skolemization, $t$ has the form $sk(W_1, \ldots, W_m)$, where $sk$ is a Skolem functor and $W_1, \ldots, W_m$ are distinct variables. Because $s$ is a ground instance of $t$, $s = t\sigma$ for some substitution $\sigma$ that grounds $t$. Because $t = sk(W_1, \ldots, W_n)$ appears in only one clause, $t$ has exactly one associated (generalized) CPT, $T$, conditional on the Skolem terms in $W_1, \ldots, W_n$. Let the parents of $s$ in $\mathcal{BN}$ be the Skolem terms in $W_1\sigma, \ldots, W_m\sigma$, and let the CPT be $T\sigma$. Note that for any node in $\mathcal{BN}$ its parents are subterms of that node. It follows that the graph structure is acyclic and hence that $\mathcal{BN}$ is a properly defined Bayes net, though possibly infinite. Therefore $\mathcal{BN}$ uniquely defines a joint distribution over ground Skolem terms; we take this to be the distribution over ground Skolem terms defined by the program $P$.

The meaning of an ordinary logic program typically is taken to be its least Herbrand model. Recall that the individuals in a Herbrand model are themselves ground terms, and every ground term denotes itself. Because we wish to consider cases where ground Skolem terms denote (non-Skolem) constants, we instead consider Herbrand quotient models [9]. In a Herbrand quotient model, the individuals are equivalence classes of ground terms, and any ground term denotes the equivalence class to which it belongs. Then two ground terms are equal according to the model if and only if they are in the same equivalence class. We take the set of minimal Herbrand quotient models for $P$ to be those derived as follows.[2] Take the least Herbrand model of the logical portion of $P$, and for each non-Skolem constant, merge zero or more ground Skolem terms into an equivalence class with that constant. This equivalence class is a new individual, replacing the merged ground terms, and it participates in exactly the relations that at least one of its members participated in, in the same manner. It follows that each resulting model also is a model of $P$. The set of models that can be constructed in this way is the set $S$ of minimal Herbrand quotient models of $P$. Let $D$ be any probability distribution over $S$ that is consistent with the distribution over ground Skolem terms defined by $P$. By consistent, we mean that for any ground Skolem term $t$ and any constant $c$, the probability that $t = c$ according to the distribution defined by $P$ is exactly the sum of the probabilities according to $D$ of the models in which $t = c$. At least one such distribution $D$ exists, since $S$ contains one model for each possible combination of equivalences. We take such $\langle D, S \rangle$ pairs to be the models of $P$.

**Agreement Between Operational and Model-theoretic Semantics**  Following ordinary logic programming terminology, the negation of a query is called the "goal," and is a clause in which every literal is negated. Given a program and a goal, the CLP($\mathcal{BN}$) operational semantics will yield a derivation of the empty clause if and only if every model $\langle D, S \rangle$ of the CLP($\mathcal{BN}$) program falsifies the goal and hence satisfies the query for some substitution to the variables in the query. This follows from the soundness and completeness of SLD-resolution. But in contrast to ordinary Prolog, the proof will be accompanied by a Bayes net whose nodes are labeled by Skolem terms appearing in the query or proof. The following theorem states that the answer to any query of this attached Bayes net will agree with the answer that would be obtained from the distribution $D$, or in other words, from the distribution over ground Skolem terms defined by the program $P$. Therefore the operational and model-theoretic semantics of CLP($\mathcal{BN}$) agree in a precise manner.

---

[1] This can be extended to a finite subset of the set of ground terms not containing Skolem symbols (functors or constants). We restrict ourselves to constants here merely to simplify the presentation.

[2] For brevity, we simply define these minimal Herbrand quotient models directly. In the full paper we show that it is sufficient to consider only Herbrand quotient models, rather than all logical models. We then define an ordering based on homomorphisms between models and prove that what we are calling the minimal models are indeed minimal with respect to this ordering.

**Theorem 1** *For any CLP($\mathcal{BN}$) program $P$, any derivation from that program, any grounding of the attached Bayes net, and any query to this ground Bayes net,[3] the answer to the query is the same as if it were asked of the joint distribution over ground Skolem terms defined by $P$.*

*Proof:* Assume there exists some program $P$, some derivation from $P$ and associated ground Bayes net $B$, and some query $Pr(q|E)$ such that the answer from $B$ is not the same as the answer from the full Bayes net $\mathcal{BN}$ defined by $P$. For every node in $B$ the parents and CPTs are the same as for that same node in $\mathcal{BN}$. Therefore there must be some path through which evidence flows to $q$ in $\mathcal{BN}$, such that evidence cannot flow through that path to $q$ in $B$. But by Lemma 2, below, this is not possible.

**Lemma 2** *Let $B$ be any grounding of any Bayes net returned with any derivation from a CLP($\mathcal{BN}$) program $P$. For every query to $B$, the paths through which evidence can flow are the same in $B$ and in the full Bayes net $\mathcal{BN}$ defined by $P$.*

*Proof:* Suppose there exists a path through which evidence can flow in $\mathcal{BN}$ but not in $B$. Consider the shortest such path; call the query node $q$ and call the evidence node $e$. The path must reach $q$ through either a parent of $q$ or a child of $q$ in $\mathcal{BN}$. Consider both cases. *Case 1*: the path goes through a parent $p$ of $q$ in $\mathcal{BN}$. Note that $p$ is a parent of $q$ in $B$ as well. Whether evidence flows through $p$ in a linear or diverging connection in $\mathcal{BN}$, $p$ cannot itself have evidence—otherwise, evidence could not flow through $p$ in $\mathcal{BN}$. Then the path from $e$ to $p$ is a shorter path through which evidence flows in $\mathcal{BN}$ but not $B$, contradicting our assumption of the shortest path. *Case 2*: the path from $e$ to $q$ flows through some child $c$ of $q$ in $\mathcal{BN}$. Evidence must flow through $c$ in either a linear or converging connection. If a linear connection, then $c$ must not have evidence; otherwise, evidence could not flow through $c$ to $q$ in a linear connection. Then the path from $e$ to $c$ is a shorter path through which evidence flows in $\mathcal{BN}$ but not $B$, again contradicting our assumption of the shortest path. Therefore, evidence must flow through $c$ in a converging connection in $\mathcal{BN}$. Hence either $c$ or one of its descendants in $\mathcal{BN}$ must have evidence; call this additional evidence node $n$. Since $n$ has evidence in the query, it must appear in $B$. Therefore its parents appear in $B$, and their parents, up to $q$. Because evidence can reach $c$ from $e$ in $B$ (otherwise, we contradict our shortest path assumption again), and a descendent of $c$ in $B$ (possibly $c$ itself) has evidence, evidence can flow through $c$ to $q$ in $B$.

---

[3]For simplicity of presentation, we assume queries of the form $Pr(q|E)$ where $q$ is one variable in the Bayes net and the evidence $E$ specifies the values of zero or more other variables in the Bayes net.

## 4 Non-determinism and Aggregates

CLP($\mathcal{BN}$) can deal with aggregates with *no extra machinery*. As an example, in Fig 1 the rating of a course depends on the satisfaction of all students who registered. Assuming, for simplicity, that the rating is a deterministic function of the average satisfaction, we must average the satisfaction Sat for all students who have a registration R in course C. The next clause shows the corresponding program:

```
rating(C, Rat) :-
    setof(S,R^(registration(R,C),
            satisfaction(R,S)), Sats),
    average(Sats, CPT),
    {Rat = rating(C) with CPT}.
```

The call to setof obtains the satisfactions of all students registered in the course. The Prolog procedure average computes the conditional probability distribution of the course's rating as a function of student satisfaction.

## 5 Recursion

Recursion provides an elegant framework for encoding sequences of events. We next show a simple example of how to encode Hidden Markov Models.

The Brukian and Arinian have lived in a state of conflict for many years. The Brukian want to send their best spy, James Bound, to spy on the Arinian headquarters. The Arinian watch commander is one of two men: Manissian is careful, but Lufy is a bit lax. The Brukian only know that if one of them was on watch today, it is likely they will be on watch tomorrow. The next program represents the information:

```
caught(0,Caught) :- !,
    {Caught = c(0) with p([t,f],[0.0,1.0],[])}.
caught(I,Caught) :-
    I1 is I-1, caught(I1, Caught0),
    watch(I, Police),
    caught(I,Caught0, Police, Caught).

watch(0, P) :- !,
    {P = p(0) with p([m,l],[0.5,0.5],[])}.
watch(I, P) :-
    I1 is I-1, watch(I1, P0),
    {P = p(I) with
        p([m,l],[0.8,0.2,0.2,0.8],[P0])}.

caught(I, C0, P, C) :-
    {C = c(I) with
```

```
p([t,f],[1.0,1.0,0.05,0.001,
         0.0,0.0,0.95,0.099],[C0,P])}.
```

The variables `c(I)` give the probability James Bound was caught at or before time `I`, and `p(I)` gives the probabilities for who is watching at time `I`.

## 6 Relationship to PRMs

Clearly from the examples in Section 2, the CLP($\mathcal{BN}$) representation owes an intellectual debt to PRMs. As the reader might suspect at this point, any PRM can be represented as a CLP($\mathcal{BN}$) program. To show this, we next present an algorithm to convert any PRM into a CLP($\mathcal{BN}$) program. Because of space limits, we necessarily assume the reader already is familiar with the terminology of PRMs.

We begin by representing the skeleton of the PRM, i.e., the database itself with (possibly) missing values. For each relational table $R$ of $n$ fields, one field of which is the key, we define $n-1$ binary predicates $r_2, \ldots, r_n$. Without loss of generality, we assume the first field is the key. For each tuple or record $\langle t_1, \ldots, t_n \rangle$ our CLP($\mathcal{BN}$) program will contain the fact $r_i(t_1, t_i)$ for all $2 \le i \le n$. If $t_i$ is a missing value in the database, then the corresponding fact in the CLP($\mathcal{BN}$) program is $r_i(t_1, sk_i)$, where $sk_i$ is a unique constant that appears nowhere else in the CLP($\mathcal{BN}$) program; in other words, $sk_i$ is a Skolem constant. It remains to represent the Bayes net structure over this skeleton and the CPTs for this structure.

For each field in the database, we construct a clause that represents the parents and the CPT for that field within the PRM. The head (consequent) of the clause has the form $r_i(Key, Field)$, where the field is the $i^{th}$ field of relational table $R$, and $Key$ and $Field$ are variables. The body of the clause is constructed in three stages, discussed in the following three paragraphs: the relational stage, the aggregation stage, and the CPT stage.

The relational stage involves generating a translation into logic of each slot-chain leading to a parent of the given field within the PRM. Recall that each step in a slot chain takes us from the key field of a relational table $R$ to another field, $i$, in that table, or vice-versa. Each such step is translated simply to the literal $r_i(X, Y)$, where $X$ is a variable that represents the key of $R$ and $Y$ is a variable that represents field $i$ of $R$, regardless of directionality. If the next step in the slot chain uses field $i$ of table $R$, then we re-use the variable $Y$; if the next step instead uses the key of table $R$ then we instead re-use variable $X$. Suppose field $i$ is the foreign key of another table $S$, and the slot chain next takes us to field $j$ of $S$. Then the slot chain is translated as $r_i(X, Y), s_j(Y, Z)$. We can use the same translation to move from field $j$ of $S$ to the key of $R$, although we would re-order the literals for efficiency. For example, suppose we are given a student key *StudentKey* and want to follow the slot chain through registration and course to find the teaching abilities of the student's professor(s). Assuming that the course key is the second field in the registration table and the student key is the third field, while the professor key is the second field of the course table, and ability is the second field of the professor table, the translation is as below. Note that we use the first literal to take us from *StudentKey* to *RegKey*, while we use the second literal to take us from *RegKey* to *CourseKey*.

$registration_3(RegKey, StudentKey),$
$registration_2(RegKey, CourseKey),$
$course_2(CourseKey, ProfKey),$
$professor_2(ProfKey, Ability)$

In the preceding example, the variable *Ability* may take several different bindings. If this variable is a parent of a field, then the PRM will specify an aggregation function over this variable, such as *mean*. Any such aggregation function can be encoded in a CLP($\mathcal{BN}$) program by a predicate definition, as in ordinary logic programming, i.e. in Prolog. We can collect all bindings for *Ability* into a list using the Prolog built-in function *findall* or *setof*, and then aggregate this list using the appropriate aggregation function such as *mean*. For the preceding example, we would use the following pair of literals to bind the variable $X$ to the mean of the abilities of the student's professors.

$findall(Ability, (\ registration_2(RegKey, CourseKey),$
$course_2(CourseKey, ProfKey),$
$professor_2(ProfKey, Ability),\ L\ ),$
$mean(L, X)$

At this point, we have constructed a clause body that will compute bindings for all the variables that correspond to parents of the field in question. It remains only to add a literal that encodes the CPT for this field given these parents.

The close link between PRMs and CLP($\mathcal{BN}$) raises the natural question, "given that we already have PRMs, of what utility is the CLP($\mathcal{BN}$) representation?" First, while there has been much work on incorporating probabilities into first-order logic, these representations have been far removed from the approach taken in PRMs. Hence while there is great interest in the relationship between PRMs and these various probabilistic logics, this relationship is difficult to characterise. Because CLP($\mathcal{BN}$) is closely related to PRMs, they may help us better understand the relationship between PRMs and various probabilistic logics. Second, because CLP($\mathcal{BN}$) programs are an extension of logic programs, they permit recursion, non-determinism (a predicate may be defined by multiple clauses), and the use of function symbols, e.g., to construct data structures such as lists or trees. We have given an example of how recursion can be used to nicely represent Hidden Markov Models. Of course we must note that the uses of recursion and recursive data structures are not unlimited. As we have seen, resolution steps that introduce a cycle into a Bayes net constraint CLP($\mathcal{BN}$), would result in cyclic terms, and thus break the declarative semantics of logic programs. Third, and most importantly from the authors' viewpoint, the following section of the paper demonstrates that the CLP($\mathcal{BN}$) representation is amenable to learning using techniques from inductive logic programming (ILP). Hence CLP($\mathcal{BN}$)s provides a way of studying the incorporation of probabilistic methods into ILP, and they may give insight into novel learning algorithms for PRMs. The methods of learning in PRMs [4] are based upon Bayes net structure learning algorithms and hence are very different from ILP algorithms. The CLP($\mathcal{BN}$) representation provides a bridge through which useful ideas from ILP might be transferred to PRMs.

## 7  Learning in CLP(BN) using ILP

This section describes the results of learning CLP($\mathcal{BN}$) programs using the ILP system ALEPH [13].

### 7.1  The School Database

We have so far used the school database as a way to explain some basic concepts in CLP($\mathcal{BN}$), relating them to PRMs. The school database also provides a good example of how to learn CLP($\mathcal{BN}$) programs.

First, we use an interpreter to to generate a sample from the CLP($\mathcal{BN}$) program. The smallest database has 16 professors, 32 courses, 256 students and 882 registrations; the numbers roughly double in each successively larger database. We have no missing data. Can we, given this sample, relearn the original CLP($\mathcal{BN}$) program?

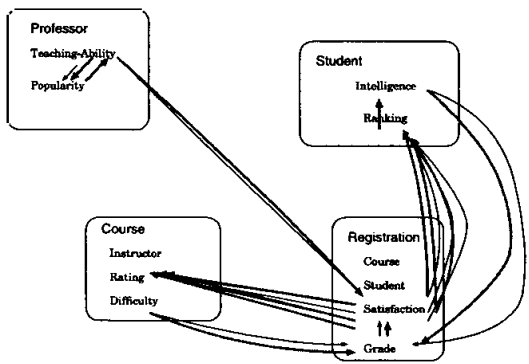

Figure 4: Pictorial representation of the CLP($\mathcal{BN}$) clauses learned from the largest schools database, before removal of cycles.

From the ILP point of view, this is an instance of multi-predicate learning. To simplify the problem we assume each predicate would be defined by a single clause. We use the Bayesian Information Criterion (BIC) score to compare alternative clauses for the same predicate. Because ALEPH learns clauses independently, cycles may appear in the resulting CLP($\mathcal{BN}$) program. We therefore augment ALEPH with a post-processing algorithm that simplifies clauses until no cycles remain; the algorithm is greedy, choosing at each step the simplification that will least affect the BIC score of the entire program.

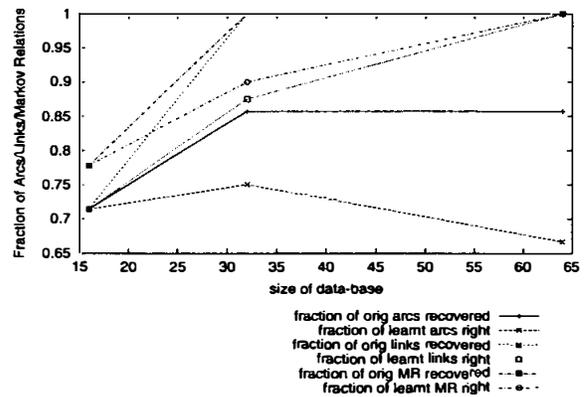

Figure 5: Graph of results of CLP($\mathcal{BN}$)-learning on the three sizes of schools databases. Links are arcs with direction ignored. A Markov relation (MR) holds between two nodes if one is in the Markov blanket of the other.

The following is one of the learned CLP($\mathcal{BN}$) clauses; to conserve space, we simply write "CPT" instead of showing the large table.

```
registration_grade(A,B) :-
    registration(A,C,D), course(C,E),
    course_difficulty(C,F), student_intelligence(D,G),
    {F = registration_grade(A) with p([a,b,c,d],CPT,[F,G]}.
```

Figure 4 illustrates, as a PRM-style graph, the full set of clauses learned for the largest of the databases before simplification; this would be the best network according to BIC, if not for the cycles. Figure 5 plots various natural measures of the match between the learned program *after cycles have been removed* and the original program, as the size of the database increases. By the time we get to the largest of the databases, the only measures of match that do not have a perfect score are those that deal with the directions of arcs.

### 7.2  Metabolic Activity

Our second application is based on data provided for the 2001 KDD Cup. We use the KDD01 Task 2 training data, consisting of 4346 entries on the activities of 862 genes [3]. We chose to study this problem because it is a real-world relational problem with much missing data. We concentrate on the two-class problem of predicting whether a gene codes for metabolism, because it illustrates a strength of learning a CLP($\mathcal{BN}$) program rather than an ordinary logic program.

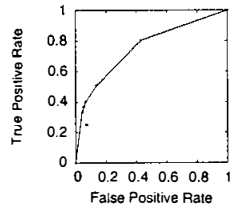

Figure 6: ROC Curve for Metabolism Application

A theory predicts metabolism if we have at least a proof saying that the probability for metabolism is above a certain value $T^M$, and no proof saying that the probability for metabolism is below a certain $T_m$. We allow $T^M$ to range from 0.3 to 0.9. Because we obtain probabilities, it is straightforward to generate an ROC curve showing how we can gain a higher true positive prediction rate if we are willing to live with a higher false positive prediction rate. This is an advantage of learning a CLP($\mathcal{BN}$) program rather than an ordinary logic program. The ROC curve is shown in Figure 6. The performance of the ordinary logic program learned by ALEPH is a single point, lying slightly (not significantly) below the curve shown.

## 8 Conclusion

This paper has defined CLP($\mathcal{BN}$), an extension of logic programming that defines a joint probability distribution over the denotations of ground Skolem terms. Because such terms are used in logic programming in place of existentially-quantified variables, or missing values, CLP($\mathcal{BN}$) is most closely related to PRMs. The full paper also gives a detailed discussion of the relationship of CLP($\mathcal{BN}$) to other probabilistic logics, including the work of Breese [2], Haddawy and Ngo [5], Sato [12], Poole [10], Koller and Pfeffer [8], Angelopolous [1], and Kersting and DeRaedt [6, 7]. To summarise that discussion in a sentence, CLP($\mathcal{BN}$) does not replicate any of these approaches because they define probability distributions over sets of objects other than the set of ground Skolem terms.

CLP($\mathcal{BN}$) is a natural extension of logic programming that subsumes PRMs. While this does not imply that CLP($\mathcal{BN}$) is preferable to PRMs, its distinctions from PRMs follow from properties of logic programming. These distinctions include the combination of function symbols and recursion, non-determinism (a predicate may be defined by multiple clauses), and the applicability of ILP techniques for learning CLP($\mathcal{BN}$) programs. Our primary direction for further work is in developing improved ILP algorithms for learning CLP($\mathcal{BN}$) programs. Other issues of interest are improving the performance of CLP($\mathcal{BN}$) through the use of tabling [11] to avoid redundant computations. Tabling would also allow CLP($\mathcal{BN}$) to support logic programs with negation.

### Acknowledgements

This work was supported by DARPA EELD grant number F30602-01-2-0571, NSF Grant 9987841 and by NLM grant NLM 1 R01 LM07050-01. Vítor Santos Costa was on leave from UFRJ and was partially supported by CNPq. We gratefully acknowledge discussions with Inês Dutra, Peter Haddawy, Kristian Kersting, and Ashwin Srinivasan.